\def\year{2020}\relax
\documentclass[letterpaper]{article} 
\usepackage{aaai20}  
\usepackage{times}  
\usepackage{helvet} 
\usepackage{courier}  
\usepackage[hyphens]{url}  
\usepackage{graphicx} 
\usepackage{graphicx}  
\usepackage{amsfonts}  
\usepackage{amsmath}
\usepackage{multirow}
\usepackage[ruled,linesnumbered]{algorithm2e}

\DeclareMathOperator{\argmin}{argmin}

\title{Posterior-Guided Neural Architecture Search}
\author{
Yizhou Zhou,\textsuperscript{\rm 1}\thanks{This work was performed while Yizhou Zhou was an intern with Microsoft Research Asia.}
Xiaoyan Sun,\textsuperscript{\rm 2}
Chong Luo,\textsuperscript{\rm 2}
Zheng-Jun Zha,\textsuperscript{\rm 1}\thanks{Corresponding author.}
Wenjun Zeng\textsuperscript{\rm 2} \\ \Large
\textsuperscript{\rm 1}University of Science Technology of China ~~~~~~~~~~~~~~~ \textsuperscript{\rm 2}Microsoft Research Asia \\
~~~~~~~~~~~zyz0205@mail.ustc.edu.cn ~~~ zhazj@ustc.edu.cn ~~~~~~~~~~~~~~ \{xysun, cluo, wezeng\}@microsoft.com 
}

\begin{document}
\maketitle

\begin{abstract}
The emergence of neural architecture search (NAS) has greatly advanced the research on network design. Recent proposals such as gradient-based methods or one-shot approaches significantly boost the efficiency of NAS. 
In this paper, we formulate the NAS problem from a Bayesian perspective. We propose explicitly estimating the joint posterior distribution over pairs of network architecture and weights. Accordingly, a hybrid network representation is presented which enables us to leverage the Variational Dropout so that the approximation of the posterior distribution becomes fully gradient-based and highly efficient. A posterior-guided sampling method is then presented to sample architecture candidates and directly make evaluations. As a Bayesian approach, our posterior-guided NAS (PGNAS) avoids tuning a number of hyper-parameters and enables a very effective architecture sampling in posterior probability space. Interestingly, it also leads to a deeper insight into the weight sharing used in the one-shot NAS and naturally alleviates the mismatch between the sampled architecture and weights caused by the weight sharing.
We validate our PGNAS method on the fundamental image classification task. Results on Cifar-10, Cifar-100 and ImageNet show that PGNAS achieves a good trade-off between precision and speed of search among NAS methods. For example, it takes 11 GPU days to search a very competitive architecture with 1.98\% and 14.28\% test errors on Cifar10 and Cifar100, respectively.
\end{abstract}

\section{Introduction}
Neural architecture search (NAS), which automates the design of artificial neural networks (ANN), has received increasing attention due to its ability of finding ANNs with similar or even better performance than manually designed ones. Essentially, NAS is a bi-level optimization problem. Given an neural architecture $\alpha$ which belongs to a pre-defined search space $G$, the lower-level objective optimizes the weight $w_{\alpha}$ of the architecture $\alpha$ as
\begin{equation}
    w^{*}_{\alpha}=\argmin_{w_{\alpha}} \mathcal{L}(\mathcal{M}(\alpha, w_{\alpha});\mathcal{D}_t),
    \label{opt_Weights}
\end{equation}
where $\mathcal{L}$ is a loss criterion that measures the performance of network $\mathcal{M}(\alpha, w_{\alpha})$ with architecture $\alpha$ and weight $w_{\alpha}$ on the training dataset $\mathcal{D}_t$; whereas 
the upper-level objective optimizes the network architecture $\alpha$ with the weight $w^{*}_\alpha$ that has been optimized by the lower-level task as
\begin{equation}
    \alpha^{*}=\argmin_{\alpha \in G} \mathcal{L}(\mathcal{M}(\alpha, w^{*}_\alpha);\mathcal{D}_v),
\end{equation}
on the validation dataset $\mathcal{D}_v$. 
To solve this bi-level problem, different approaches have been proposed, e.g. evolutionary-based methods  \cite{liu2017hierarchical,real2018regularized}, reinforcement learning based schemes \cite{baker2016designing,zoph2016neural,zoph2018learning,tan2018mnasnet,zhong2018practical,zela2018towards,real2018regularized,baker2017accelerating,swersky2014freeze,domhan2015speeding,klein2016learning,liu2018progressive} or gradient-based algorithms \cite{liu2018darts,cai2018proxylessnas,brock2017smash,xie2018snas}. However, most of these methods suffer from high computational complexity (often in the orders of thousands of GPU days) \cite{liu2017hierarchical} \cite{real2018regularized} \cite{baker2016designing} \cite{zoph2016neural} \cite{zoph2018learning}, or lack of convergence guarantee \cite{cai2018proxylessnas,liu2018darts,xie2018snas}.



Instead of directly tackling the bi-level optimization problem, some attempts \cite{Wu2018FBNetHE,saxena2016convolutional,shin2018differentiable,xie2018snas} 
relax the discrete search space $G$ to be continuous. Given one continuous relaxation $G_r$ of topology $r$, 
the weight and architecture can be jointly optimized by the single objective function
\begin{equation} 
    r^{*},w^{*} = \argmin_{r,w} \mathcal{L}(\mathcal{M}({G}_{r},w);\mathcal{D}_t).
    \label{Diff}
\end{equation}
Then the optimal architecture $\alpha^{*}$ is derived by discretizing the continuous one $G_{r^{*}}$. These methods greatly simplify the optimization problem and enable end-to-end training.
However, since the validation set $\mathcal{D}_v$ is excluded in Eq. \ref{Diff}, the search results are usually biased on training datasets.

More recent NAS methods tend to reduce the computational complexity by decoupling the bi-level optimization problem into a sequential one \cite{bender2018understanding,brock2017smash,guo2019single}. Specifically, the search space $G$ here is defined by an over-parameterized super-network (one-shot model) with architecture $\alpha_o$. 
Then the one-shot NAS starts with 
optimizing the weights $w_{\alpha_o}$ of the super-network $\alpha_o$ by Eq. \ref{opt_Weights}, resulting $w_{\alpha_o}^{*}$ as
\begin{equation}\label{oneshotstep1}
     w_{\alpha_o}^{*} = \argmin_{w_{\alpha_o} }\mathcal{L}(\mathcal{M}(\alpha_o,w_{\alpha_o});\mathcal{D}_t).
\end{equation}
After that, a number of sub-architectures {$\alpha$} are sampled from $\alpha_o$. Then the best-performing one is selected by
\begin{equation} \label{oneshotstep2}
    \alpha^{*} = \argmin_{\alpha \subseteq \alpha_o}\mathcal{L}({\mathcal{M}(\alpha, w^{*}_{\alpha}});\mathcal{D}_v),
\end{equation}
where $w^{*}_{\alpha}$ is inherited from $w^{*}_{\alpha_o}$.
We notice that one core assumption in the one-shot NAS method is that the best-performing sub-network shares weights with the optimal super-network. Thus the sampled sub-networks do not need to be re-trained in the searching process. This assumption, on the one hand, greatly boosts the efficiency of NAS; on the other, it could lead to mismatches between weights and architectures of sampled sub-networks, which jeopardizes the following ranking results \cite{xie2018snas}. More clues can be found that most one-shot methods rely on fine-tuning to further improve the performance of the found model. In addition, the sampling process in current one-shot NAS is much less explored which has a big impact on the performance as demonstrated later in our study (Table \ref{ablationx_a}). 


In this paper, we propose modeling the NAS problem from a Bayesian perspective and accordingly present a new NAS strategy, \textit{i.e.} posterior-guided NAS (PGNAS). In PGNAS, given a search space $G$, we estimate the posterior distribution $p(\alpha,w \mid \mathcal{D}_{t})$ ($\alpha \in G $) over pairs of architecture and weight ($\alpha,w$). Then optimal architecture is searched by 
\begin{equation}\label{intr_smpl}
    \alpha^{*} = \argmin_{\alpha, w \sim p{(\alpha,w \mid \mathcal{D}_t)}} \mathcal{L}(\mathcal{M}(\alpha, w);\mathcal{D}_v).
\end{equation}
However, since the posterior distribution $p(\alpha,w \mid \mathcal{D}_{t})$ is intractable, we approximate it by 
a variational distribution $q_{\theta}(\alpha,w)$ as 
\begin{equation}\label{intr_poster}
    \theta^* = \argmin_{\theta} L (q_{\theta}(\alpha,w), p(\alpha,w \mid \mathcal{D}_{t})),
\end{equation}
where $\theta$ denotes the variational parameters and $L$ measures the distance between two distributions. 
Still, finding $\theta^*$ is not trivial. Therefore, we further propose a hybrid network representation to facilitate an end-to-end 
trainable solution.
In short, our PGNAS manages to leverage the training dataset to estimate the joint posterior distribution over network architecture and weight, based on which an efficient and effective sampling in posterior probability space is enabled. 

We recently noticed that there is a parallel work \cite{zhou2019bayesnas} that employs the sparse Bayesian learning (SBL) \cite{tipping2001sparse} in NAS. This work focuses on tackling two problems in the one-shot NAS, i.e., neglect of dependencies between nodes and problematic magnitude-based operation pruning. It encodes the node dependency logic into the prior distribution of the architecture and exploit the SBL
to obtain the most sparse solution for the prior distribution. The entropy of the derived prior distribution with a pre-defined threshold is then used as the criterion to prune operations. Differently, we directly formulate NAS in a posterior probability space, and sample pairs of architecture and weights in this space to search for good architectures.

In summary, the main contributions of this work are:
\begin{itemize}
    \item We convert NAS to a distribution construction problem. We formulate the NAS problem by Bayesian theory and propose to estimate the joint posterior distribution of pairs of network architecture and weights. It thus enables a very efficient search process in the probability space through a posterior-guided sampling. The presented PGNAS achieves a good trade-off between performance and speed of search, e.g. it speeds up the search process by 20X compared with the second best while achieving the best performance with 1.98\% test error on Cifar10.
    \item An end-to-end trainable solution is proposed to approximate the joint posterior distribution on architecture and weights. In particular, a hybrid network representation is presented to facilitate the Variational Dropout \cite{gal2016dropout}-based solution, which makes the approximation fully gradient-based and highly efficient.
    \item Our PGNAS models and samples architecture and weights jointly. It thus reduces the mismatch between the architecture and weights caused by weight sharing in the one-shot NAS and improves the reliability of evaluation for the sampled network. In our PGNAS, the sampled weights can be adopted directly by the searched architecture to achieve high performance without fine-tuning.
    \item We find that the weight sharing in our PGNAS can be viewed as a re-parametrization that enables us to estimate the posterior distribution via end-to-end optimization. This finding may help in better understanding of weight sharing that is well-accepted in the one-shot NAS.
\end{itemize}

\section{PGNAS}

In this section, we first formulate the target problem in our PGNAS via the joint posterior distribution. Then an end-to-end trainable solution is presented to approximate the posterior distribution. 
At last, an efficient sampling and ranking scheme is described to facilitate the search process. 

In the following, 
we use $w_l^s \in R^{K_l \times H_l \times s \times s}$ 
to denote the convolution weight matrix in layer $l$ with spatial kernel size $s$, where $K_l$ and $H_l$ denote  the number of input and output channels in layer $l$, respectively.  $w_{l,k}^s \in R^{1 \times H_l \times s \times s}$ denotes the sliced kernel operated on the $k^{th}$ ($1\leq k\leq K_l$) input channel dimension. $w=\{w_{l,k}^s\}$ presents weights of the whole super-network. As deriving an architecture $\alpha$ is equivalent to selecting a set of convolution kernels, we introduce a set of random variables $\boldsymbol{\alpha}=\{\boldsymbol{\alpha}_{l,k}^s\}$ ( $\boldsymbol{\alpha}_{l,k}^s\ \in \{0, 1\}$) to indicates deactivating or activating convolution kernel $\boldsymbol{w}_{l,k}^s$ by setting the corresponding $\boldsymbol{\alpha}_{l,k}^s$ to 1 or 0, respectively. We use boldface for random variables later on.

\subsection{Problem Formulation}

As defined in Eq. \ref{intr_smpl} and \ref{intr_poster}, in order to search a good architecture from the posterior probability space, we need to construct a joint posterior distribution over $\boldsymbol{\alpha}$ and $\boldsymbol{w}$ which is usually very hard to obtain. However, we notice it is not necessary to compute it explicitly as deactivating or activating a convolution kernel is also equivalent to multiplying a binary mask to the kernel. Hence, we combine the two random variables into a new one $\boldsymbol{\varphi} = \{\mathcal{\varphi}_{l,k}^s\}$, where $\mathcal{\varphi}_{l,k}^s\ = \boldsymbol{w}_{l,k}^s \cdot \boldsymbol{\alpha}_{l,k}^s$. Now, the key problem in our PGNAS becomes the approximation of the posterior distribution over the \emph{hybrid network representation} $\boldsymbol{\varphi}$. Mathematically,
\begin{equation}\label{unified_representation}
\begin{split}
     p(\boldsymbol{\varphi} \mid X, Y) 
     = \dfrac{p(Y \mid X, \boldsymbol{\varphi})p(\boldsymbol{\varphi})}{\int_{\boldsymbol{\varphi}} p(Y \mid X, \boldsymbol{\varphi})}, 
\end{split}
\end{equation}
where $X=\{x_i mid i=1,...,N\}$ and $Y=\{y_i \mid i=1,...,N\}$ denote $N$ training samples and labels, respectively. $p(Y \mid X, \boldsymbol{\varphi})$ is the likelihood that can be inferred by $\prod_{i=1}^N p(y_i \mid f^{\boldsymbol{\varphi}}(x_i))$, where $f^{\boldsymbol{\varphi}}$ denotes a sub-network defined by hybrid representation $\boldsymbol{\varphi}$. $p(\boldsymbol{\varphi})$ is the a priori distribution of hybrid representation $\boldsymbol{\varphi}$. Because the marginalized likelihood $\int_{\boldsymbol{\varphi}} p(Y \mid X, \boldsymbol{\varphi})$ in Eq. \ref{unified_representation} is intractable, we use a variational distribution $q_{\theta}(\boldsymbol{\varphi})$ to approximate Eq. \ref{unified_representation} and reformulate our target problem as
\begin{equation}\label{objrewrt}
\begin{split}
    &\theta^* = \argmin_{\theta} L_d (q_{\theta}(\boldsymbol{\varphi}), p(\boldsymbol{\varphi} \mid \mathcal{D}_{t})),\\
    &\alpha^{*} = \argmin_{\varphi \sim q_{\theta^*}(\boldsymbol{\varphi})} \mathcal{L}(\varphi;D_v).
\end{split}
\end{equation}
Here we choose KL divergence and accuracy to instantiate $\mathcal{L}_d$ and $\mathcal{L}$, respectively.

\subsection{Posterior Distribution Approximation} \label{APosterioriApproximation}
To solve Eq. \ref{objrewrt}, we employ Variational Inference(VI) which minimizes the negative Evidence Lower Bound (ELBO)
\begin{equation}\label{VI}
\begin{split}
    \mathcal{L}_{VI}(\theta) :&= KL(q_\theta(\boldsymbol{\varphi}) \mid\mid p(\boldsymbol{\varphi})) \\
    &- \sum_{i=1}^{N} \int q_\theta(\boldsymbol{\varphi}) \log p(y_i \mid f^{\boldsymbol{\varphi}}(x_i))d\boldsymbol{\varphi},    
\end{split}
\end{equation}
Here we propose solving Eq. \ref{VI} by the network friendly Variational Dropout.
\begin{algorithm}
\small
\caption{PGNAS}\label{PGNAS_alg}
\KwData{Training dataset $D_t$, validation dataset $D_v=\{X, Y\}$, learning rate $\eta$, number of candidates C}
\KwResult{Searched architecture $\alpha^*$}

\While{Not Converged}
{
Sample M pairs of training sample (x, y) from $D_t$ \; 
Sample M random variables $\epsilon_i = \{\boldsymbol{\epsilon}_{l,k}^s\}$, where $\boldsymbol{\epsilon}_{l,k}^s \sim Bernoulli(p^{s}_l)$ \;
$p^{s}_l \leftarrow p^{s}_l + \frac{\partial}{\partial p^{s}_l} [-\frac{1}{M}\sum_{i=1}^{M}p(y_i \mid f^{\theta, \boldsymbol{\epsilon}_i}(x_i))] + \frac{\partial}{\partial p^{s}_l}[(\sum_{l,k,s} \frac{(l_{l,k}^s)^{2}(1-p_l^s)}{2N} \Arrowvert m_{l,k}^s \Arrowvert^2) + \frac{1}{N}\mathcal{H}(p^{s}_l)]$ \;

$m^{s}_{l,k} \leftarrow m^{s}_{l,k} + \frac{\partial}{\partial m^{s}_{l,k}} [-\frac{1}{M}\sum_{i=1}^{M}p(y_i \mid f^{\theta, \boldsymbol{\epsilon}_i}(x_i))] + \frac{\partial}{\partial m^{s}_{l,k}}[(\sum_{l,k,s} \frac{(l_{l,k}^s)^{2}(1-p_l^s)}{2N} \Arrowvert m_{l,k}^s \Arrowvert^2)]$ \;
}

Sample C random variables $\epsilon_i = \{\boldsymbol{\epsilon}_{l,k}^s\}$, where $\boldsymbol{\epsilon}_{l,k}^s \sim Bernoulli(p^{s}_l)$ \;
Initialize $\gamma_{best}$ \;
\For{each $\epsilon_i$}
{
Compute performance $\gamma_i$ on $D_v$ based on $p(Y \mid f^{\theta, \epsilon_i}(X))$ \;
\If{$\gamma_i$ is better than $\gamma_{best}$}
{
$\epsilon_{best} \leftarrow \epsilon_i$ \;
$\gamma_{best} \leftarrow \gamma_{i}$
}
}
Derive $\alpha^*$ from $\epsilon_{best}$
\end{algorithm}
\subsubsection{Approximation by Network Training}

Due to the hybrid network representation, we can use the re-parametrization trick \cite{kingma2013auto} as in \cite{gal2016uncertainty,gal2016dropout} to solve Eq. \ref{VI}. We choose a deterministic and differentiable transformation function $g(\cdot,\cdot)$ that re-parameterizes the $q_\theta(\boldsymbol{\varphi})$ as $\boldsymbol{\varphi}=g(\theta, \boldsymbol{\epsilon})$, where $\boldsymbol{\epsilon} \sim p(\boldsymbol{\epsilon})$ is a parameter-free distribution. Take a uni-variate Gaussian distribution $x \sim q_\theta(x)=\mathcal{N}(\mu, \sigma)$ as an example, its re-parametrization can be $x = g(\theta, \boldsymbol{\epsilon}) = \mu + \sigma \boldsymbol{\epsilon}$ with $\boldsymbol{\epsilon} \sim \mathcal{N}(0,1)$, where $\mu$ and $\sigma$ are the variational parameters $\theta$. Gal \textit{et.al.} in \cite{gal2016uncertainty,gal2016dropout} show that when the network weight is re-parameterized with 
\begin{equation}\label{repara_omega}
\begin{split}
    \boldsymbol{w}_{l,k}^s = m_{l,k}^s \cdot \boldsymbol{z}_{l,k}^s,~where~\boldsymbol{z}_{l,k}^s \sim Bernoulli(\widetilde{p}_{l}^s),
\end{split}
\end{equation}
the function draw w.r.t. variational distribution over network weights $\boldsymbol{w}$ can be efficiently implemented via network inference. Concretely, the function draw is equivalent to randomly drawing masked deterministic weight matrix $m = \{m^s_{l,k}\}$ in neural networks, which is known as the Dropout operations \cite{srivastava2014dropout}. Similarly, we replace $\boldsymbol{w}_{l,k}^s$ in our hybrid representation $\boldsymbol{\varphi}_{l,k}^s = \boldsymbol{w}_{l,k}^s \cdot \boldsymbol{\alpha}_{l,k}^s$ with $m_{l,k}^s \cdot \boldsymbol{z}_{l,k}^s$, and reformulate $\boldsymbol{\varphi}_{l,k}^s$ as
\begin{equation}\label{reparam-1}
    \begin{split}
        \boldsymbol{\varphi}_{l,k}^s\ = m_{l,k}^s \cdot \boldsymbol{\epsilon}_{l,k}^s, ~where~\boldsymbol{\epsilon}_{l,k}^s=\boldsymbol{z}_{l,k}^s \cdot \boldsymbol{\alpha}_{l,k}^s,
    \end{split}
\end{equation}
In Eq. \ref{reparam-1}, we have an additional random variable $\boldsymbol{\alpha}_{l,k}^s$ that controls the activation of kernels whose distribution is unknown. Here we propose using the marginal probability $p(\boldsymbol{\alpha}_{l,k}^s \mid X, Y)$ to characterize its behavior, because the marginal can reflect the expected probability of selecting kernel $\boldsymbol{\alpha}_{l,k}^s$ given the training dataset. It exactly matches the real behavior if the selections of kernels in a super-network are independent. 
Since the joint distribution of network architecture $\boldsymbol{\alpha}=\{\boldsymbol{\alpha}_{l,k}^s\}$ is a \textit{multivariate Bernoulli distribution}, its marginal distribution obeys $Bernoulli(\overline{p}_{l,k}^s)$ \cite{dai2013multivariate}, where $\overline{p}_{l,k}^s$ is to be optimized. Therefore, we have
\begin{equation}\label{repara_phi}
\begin{split}
    \boldsymbol{\varphi}_{l,k}^s\ = m_{l,k}^s \cdot \boldsymbol{\epsilon}_{l,k}^s,
    ~where~\boldsymbol{\epsilon}_{l,k}^s \sim Bernoulli(\widetilde{p}_{l}^s \cdot \overline{p}_{l}^s).
\end{split}
\end{equation}
 Here we omit the subscript $k$ in the original $Bernoulli(\overline{p}_{l,k}^s)$ because the importance of branches which come from the same kernel size group and layer should be identical. By replacing $\widetilde{p}_{l}^s \cdot \overline{p}_{l}^s$ with a new variable $p_{l}^s$, Eq. \ref{repara_phi} has the same form as Eq. \ref{repara_omega}. Then, we can obtain
\begin{equation}\label{MC_refm}
    \begin{split}
       \mathcal{L}_{MC}(\theta) := & KL(q_\theta(\boldsymbol{\varphi}) \mid\mid p(\boldsymbol{\varphi})) - \sum_{i=1}^{N} \log p(y_i \mid f^{\theta, \boldsymbol{\epsilon}_i}(x_i)), \\
    &s.t.~\mathbb{E}_{\boldsymbol{\epsilon}}\{\mathcal{L}_{MC}(\theta)\} = \mathcal{L}_{VI}(\theta).
    \end{split}
\end{equation}
where variational parameters $\theta=\{m_{l,k}^s\}$ are composed of the deterministic kernel weights. $\boldsymbol{\epsilon}_i=\{\boldsymbol{\epsilon}_{l,k}^s\}_i$ are the $i^{th}$ sampled random variables which encodes the distribution of network architecture.
Eq. \ref{MC_refm} indicates that the (negative) ELBO can be computed very efficiently. It is equivalent to the KL term minus the log likelihood that is inferenced by the super-network $f^{\boldsymbol{\varphi}}$ (now reparameterized as $f^{\theta, \boldsymbol{\epsilon}}$). During each network inference, convolution kernels are randomly deactivated w.r.t. probability $p=\{p_l^s\}$, which is exactly equivalent to a dropout neural network. 

Now, approximating posterior distribution over the hybrid network representation is converted to optimizing the network $f^{\boldsymbol{\varphi}}$ with dropout and a KL regularization term. If the derivatives of both terms are tractable, we can efficiently optimize it in an end-to-end fashion.

\subsubsection{Network Optimization}\label{optimization}
In addition to the variational parameters $\theta$, the variable $\widetilde{p}$ and $\overline{p}$ in Eq. \ref{repara_phi} should also be optimized (either via grid-search \cite{gal2016uncertainty} or gradient-based method \cite{gal2017concrete}). So we need to compute $\frac{\partial}{\partial p \partial m}\mathcal{L}_{MC}(\theta)$. If each convolution kernel is deactivated with a prior probability $u^s_{l,k}$ along with a Gaussian weight prior $\mathcal{N}(\boldsymbol{w}_{l,k}^s;0,I/(d_{l,k}^s)^{2})$, then the priori distribution for the hybrid representation $\boldsymbol{\varphi}$ is exactly a \textit{spike and slab} prior $p(\boldsymbol{\varphi}_{l,k}^s)=u^s_{l,k} \cdot \delta(\boldsymbol{w}_{l,k}^s-0)+(1-u^s_{l,k}) \cdot \mathcal{N}(\boldsymbol{w}_{l,k}^s;0,I/(d_{l,k}^s)^{2})$, where $d^s_{k,l}$ is prior length scale. Following \cite{gal2016dropout,gal2017concrete}, the derivatives of Eq. \ref{MC_refm} can be approximated as
\begin{equation}\label{deriv}
    \begin{split}
        \frac{\partial}{\partial p^{s}_l \partial m^{s}_{l,k}} [&-\frac{1}{N}(\sum_{i=1}^{N}p(y_i \mid f^{\theta, \boldsymbol{\epsilon}_i}(x_i)) + \mathcal{H}(p^{s}_l))\\
         &+\sum_{l,k,s} \frac{(d_{l,k}^s)^{2}(1-p_l^s)}{2N} \Arrowvert m_{l,k}^s \Arrowvert^2],
    \end{split}
\end{equation}
where $\mathcal{H}(p^{s}_l)=\sum_{l,s}k_l^s \cdot p_{l}^s \cdot \log p_l^s$ and $k_l^s$ denotes the number of input channels for convolution kernel of spatial size $s$ at layer $l$. Please note that the above derivation is obtained by setting the prior $u$ to be zero, which indicates the network architecture prior is set to be the whole super-network. The motivation of employing $u=0$ is that a proper architecture prior is usually difficult to acquire or even estimate, but $u=0$ can be a reasonable one when we choose the over-parameterized network that proves effective on many tasks as our super-network. Besides, $u=0$ provides us a more stable way to optimize the $\mathcal{L}_{MC}(\theta)$ \cite{gal2016uncertainty}. So, we will use the super-network that are built upon manually designed networks in our experiments.

\begin{table*}
\centering
\small
\begin{tabular}{c c c c c}
\hline
\hline
Method & Error(\%) & GPUs Days & Params(M) & Search Method \\
\hline
\hline




shake-shake + cutout \cite{devries2017improved} & 2.56 & - & 26.2 & - \\

\hline


NAS + more filters \cite{zoph2016neural} & 3.65 & 22400 & 37.4 & RL \\


NASNET-A + cutout \cite{zoph2018learning} & 2.65 & 1800 & 3.3 & RL \\



PathLevel EAS + PyramidNet + cutout \cite{cai2018path} & 2.30 & 8.3 & 13.0 & RL \\


ENAS + cutout \cite{pham2018efficient} & 2.89 & 0.5 & 4.6 & RL \\

EAS (DenseNet) \cite{cai2018efficient} & 3.44 & 10 & 10.7 & RL \\
\hline

AmoebaNet-A + cutout \cite{real2018regularized} & 3.34 & 3150 & 3.2 & evolution \\


Hierachical Evo \cite{liu2017hierarchical} & 3.63 & 300 & 61.3 & evolution \\

PNAS \cite{liu2018progressive} & 3.63 & 225 & 3.2 & SMBO \\
\hline


BayesNAS + PyramidNet\cite{zhou2019bayesnas} + cutout & 2.40 & 0.1 & 3.4 & gradient-based \\

DARTS + cutout \cite{liu2018darts} & 2.83 & 4 & 3.4 & gradient-based \\

SNAS + cutout \cite{xie2018snas} & 2.85 & 1.5 & 2.8 & gradient-based \\

NAONet + cutout \cite{luo2018neural} & 2.07 & 200 & 128 & gradient-based \\
\hline

One-Shot Top \cite{bender2018understanding} & 3.70 & - & 45.3 & sampling-based \\

SMASH \cite{brock2017smash} & 4.03 & 1.5 & 16.0 & sampling-based \\
\hline

PGNAS-MI(ours)  &2.06 & 6.5 & 33.4 & guided sampling \\

PGNAS-MI$^*$(ours) & \textbf{1.98} & 11.1 & 32.8 & guided sampling \\

\hline
\hline
\end{tabular}
\caption{Performance comparison with other state-of-the-art results. Please note that we do not fine-tune the network searched by our method. $^*$ indicates the architecture searched by sampling 10000 candidates. Full table is in supplementary material.}
\label{cifar10}
\end{table*}

Intuitively, the derived Eq. \ref{deriv} tries to find a distribution that can interpret the data well (the first likelihood term) and keep the architecture as sparse as possible (the second and third regularization terms). Please note here that the L2-like term in Eq. \ref{deriv} is not a conventional L2 regularization term. Its coefficient $(d_{l,k}^s)^{2}(1-p_l^s)$ correlates with architecture selection parameter $p$ and thus encourages the selection of kernels whose learned weights $w$ are not only representative but also \textit{sparse} (as the gradients w.r.t. $p$ here rely on the sparsity of learned $w$). It is consistent with our design where we correlate the network weights and architecture by proposing the hybrid network representation $\boldsymbol{\varphi}$.

Since the first term in Eq. \ref{deriv} involves the derivatives of the non-differentiable Bernoulli distribution (remember $\boldsymbol{\epsilon}_{l,k}^s \sim Bernoulli({p}_{l}^s)$ in Eq. \ref{repara_phi}), we thus employ the Gumbel-softmax \cite{jang2016categorical} to relax the discrete distribution $ Bernoulli(p_{l}^s)$ to continuous space and the $\boldsymbol{\epsilon}$ in Eq. \ref{deriv} and Eq. \ref{repara_phi} can be deterministically drawn by
\begin{equation}\label{gumbelsft}
    \begin{split}
        \boldsymbol{\epsilon}_{l,k}^s = \sigma(\frac{1}{\tau}[&\log p_{l}^s - \log (1-p_{l}^s)
        + \log(\log r_2) \\
        & - \log(\log r_1)]) \\
        &s.t.~r_1,r_2 \sim Uniform(0,1),
    \end{split}
\end{equation}
where $\sigma$ denotes the sigmoid function and $\tau$ is the temperature that decides how steep the sigmoid function is. If $\tau$ goes to infinite, the above parametrisation is equivalent to drawing the sample from Bernoulli distribution. (Similar relaxation is used in \cite{gal2017concrete} with another re-parametrisation method)
By adopting Eq. \ref{gumbelsft}, the derivatives in Eq. \ref{deriv} can be computed. Combining the Eq. \ref{unified_representation}, \ref{VI} and \ref{MC_refm} , one can see that the posterior distribution over the hybrid representation $\boldsymbol{\varphi}$ can be approximated by simply training the super-network in an end-to-end fashion with two additional regularization terms and dropout ratio $p$.

\subsection{Sampling and Ranking} \label{SamplingAndRanking}
Once the variational distribution $q_{\theta}(\boldsymbol{\varphi})$ is obtained, we sample a group of network candidates $S = \{s_1, s_2, ..., s_C\}$ w.r.t. $q_{\theta}(\boldsymbol{\varphi})$, where the $C$ is the number of samples. According to Eq. \ref{repara_phi}, our sampling process is performed by 
activating kernels stochastically with the learned $p_l^s$, which is equivalent to regular dropout operation. Specifically, 
each candidate is sampled by randomly dropping convolution kernel $w_{l,k}^s$ w.r.t. the probability $p_l^s$ for every $l$, $k$ and $s$ in the super-network model.
Then the sampled candidates are evaluated and ranked on a held-out validation dataset. 
Due to the hybrid network representation, we actually sample architecture-weight pairs, which relieves the mismatch problem. 
At last, the best-performing one is selected by Eq. \ref{intr_smpl}. 

We summarize the complete working flow in Algorithm. \ref{PGNAS_alg} and provide proof details in supplementary materials for better understanding. Please note that the proposed PGNAS, though not intentionally, leads to an adaptive dropout that reflects the importance of different parts in the super-network. It thus relieves the dependency on the hyper-parameter sensitive, carefully designed drop-out probability in the previous one-shot methods \cite{bender2018understanding}.



\section{Experiments}

To fully investigate the behavior of the PGNAS, we test our PGNAS on six super-networks. Because we use $u=0$ to facilitate Eq. \ref{deriv}, we construct the super-networks based on architecture priors perceived from manually designed networks. We evaluate the performance of our PGNAS on three databases
Cifar-10, Cifar-100 and ImageNet, respectively.
For every super-network, we insert a dropout layer after each convolution layer according to 
Eq. \ref{gumbelsft} to facilitate the computation of Eq. \ref{deriv}. This modification introduces parameters and FLOPS of negligible overheads. Our PGNAS is trained in an end-to-end way with the Stochastic Gradient Descent (SGD) using a single P40 GPU card for Cifar-10/Cifar-100 and 4 M40 GPU cards for ImageNet. Once a model converges, we sample different convolution kernels w.r.t. the learned dropout ratio to get 1500/5000/1500 candidate architectures for Cifar-10, Cifar-100 and ImageNet, respectively. These 1500 candidates are ranked on a held-out validation dataset and the one with the best performance will be selected as the final search result. 

\subsection{Cifar-10 and Cifar-100}
~~\textbf{Super-network and Hyper-parameters.}  We test our PGNAS on Cifar-10 and Cifar-100 with the super-network, i.e. SupNet-MI, which are based on the manually designed multi-branch ResNet \cite{gastaldi2017shake}.
Please refer to the supplementary material for more details of the super-networks and all hyper-parameter settings used in this paper.

\begin{table*}
\centering
\small
\begin{tabular}{c c c c c}
\hline
\hline
Method & Error(\%) & GPUs Days & Params(M) & Search Method \\
\hline
\hline

NASNET-A \cite{zoph2018learning} & 19.70 & 1800 & 3.3 & RL \\

ENAS \cite{pham2018efficient} & 19.43 & 0.5 & 4.6 & RL \\

AmoebaNet-B \cite{real2018regularized} & 17.66 & 3150 & 2.8 & evolution \\

PNAS \cite{liu2018progressive} & 19.53 & 150 & 3.2 & SMBO \\

NAONet + cutout \cite{luo2018neural} & 14.36 & 200 & 128 & gradient-based \\
\hline
PGNAS-MI + constant L2 term(ours) & 17.41 & - & 39.6 & - \\
PGNAS-MI(ours) & \textbf{14.28} & 11 & 46.4 & guided sampling \\

\hline
\hline
\end{tabular}
\caption{State-of-art results on Cifar-100. Please note that we do not fine-tune the network searched by our method. "PGNAS-MI + constant L2 term" indicates we replace the second L2 term in Eq. \ref{deriv} with conventional a L2 term with a constant weight.}
\label{cifar100}
\end{table*}
\begin{table*}
\centering
\small
    \begin{tabular}{c| c c| c c| c c| c c }
            \hline
            \hline
            \multirow{2}{*}{} &
              \multicolumn{2}{c}{SupNet-EI} &
              \multicolumn{2}{c}{SupNet-E} &
              \multicolumn{2}{c}{SupNet-MI} & 
              \multicolumn{2}{c}{SupNet-M}  \\
            & Err. & Param. & Err. & Param. & Err. & Param. & Err. & Param. \\
            \hline
            \hline
            Full model & 2.78\% & 15.3M & 2.98\% & 4.6M & -\% & 72.7M & 2.58\% & 26.2M\\
            \hline
            Random w/o FT & 13.45\% & 10.7M & 15.87\% & 3.0M & 9.75\% & 35.4M & 2.63\% & 22.4M \\
            \hline
            Random w/ FT & 3.16\% & 10.7M & 3.47\% & 3.0M & 2.69\% & 35.4M & 2.56\% & 22.4M \\
            \hline
             PGNAS & \textbf{2.56}\% & 10.8M & \textbf{2.73}\% & \textbf{3.1}M & \textbf{2.06}\% & \textbf{33.4}M & \textbf{2.20}\% & \textbf{21.6}M \\
            \hline
            \hline
    \end{tabular}
\caption{Impact of the guided sampling. w/o FT and w/ FT indicate whether the searched one is fine-tuned on the dataset.}
\label{ablationx_a}
\end{table*}

\begin{table}[]
\centering
     \begin{tabular}{c|ccccc}
               $l^2$  & 50 & 150 & 250 & 500 \\
                 \hline
        Error(\%) & 2.13 & \textbf{2.06} & 2.27 & 2.39 \\
        Params(M) & 49.9 & 33.4 & 23.8 & 18.2 \\
        \hline
    \end{tabular}
\caption{Impact of the weight prior $l^2$ on SupNet-EI.}
\label{ablationx_b}
\end{table}

\begin{table}[]
\centering
    \begin{tabular}{c|ccc}
                 & EI & M & EI$^{\dagger}$ \\
                 \hline
        $\tau=\frac{2}{3}$  & 2.74\%         & 2.49\%         & 2.68\%      \\
        $\tau=\frac{1}{5}$ & 2.56\%         & 2.20\%          & -      \\
        \hline
    \end{tabular}
\caption{Impact of the temperature $\tau$. $^\dagger$ denotes fine-tuning.}
\label{ablationx_c}
\end{table}

\textbf{Comparison with State-of-the-arts.} Table. \ref{cifar10} shows the comparison results on Cifar-10. Here PGNAS-X denotes the performance of our PGNAS on the super-network SupNet-X. From top to bottom, the first group consists of state-of-the-art manually designed architectures on Cifar-10; the following three groups list the related NAS methods utilizing different algorithms, e.g. RL, evolution, and gradient decent; the last group exhibits the performance of PGNAS. 

Please note that the search spaces utilized in each work are quite different. For example, \cite{zoph2018learning,real2018regularized,liu2018darts,pham2018efficient} employ cell-based search space where each cell contains 4 nodes and there are 5-11 operation candidates between two nodes, \cite{cai2018efficient,cai2018path,zhou2019bayesnas} utilize DenseNet and PyramidNet \cite{han2017deep} as base network, respectively, and \cite{luo2018neural,bender2018understanding,brock2017smash}  apply search algorithm on their self-designed search space. PGNAS-MI incorporates the multi-branch ResNet which only provides up to 4 operation candidates between two nodes (layers) as the initial model. It suggests that our search space is much smaller than the one used in state-of-the-art works such as NAONet, EAS and PathLevel EAS. Still, the proposed PGNAS is capable of finding very advanced architectures in a efficient and effective way, e.g. it finds the architecture at the lowest errors 1.98\% on 11.1 GPU days only.

We also enlist the multi-branch ResNet \cite{gastaldi2017shake} that inspires the design of our super-network in Table \ref{cifar10}. Our PGNAS-MI outperform "shake-shake+cutout" by 0.58\%. Regarding the sampling based one-shot method "One-Shot Top" which achieves a 3.7\% classification error by randomly sampling 20000 architectures, our PGNAS attains a much higher performance by sampling only 1500 network architectures due to the posterior distribution guided sampling. 

Table. \ref{cifar100} further demonstrate the performance of our PGNAS on a much challenging dataset Cifar-100. 
Our PGNAS achieves a good trade-off on efficiency and accuracy. It achieves 14.28\% error rate with only 11 GPU days, which outperforms the most advanced results NAONet in terms of both model performance and search time.

Please note that results of our PGNAS are achieved during search process without any additional fine-tuning on weights of the searched architectures, while those of other methods are obtained by fine-tuning the searched models. In the following ablation study, we will discuss more on this point.

\textbf{Ablation Study and Parameter Analysis.}
We first evaluate the effect of our posterior distribution guided sampling method in Table. \ref{ablationx_a}. In order to demonstrate the generalization of PGNAS, in addition to the SupNet-MI/SupNet-M which are based on the multi-branch ResNet,
we also apply PGNAS to the architectures obtained by ENAS \cite{pham2018efficient}. We denote them as SupNet-EI/SupNet-E.  Please refer to the supplementary materials for more details.

Compared with the baseline "Random" sampling that is implemented by employing predefined dropout strategy as discussed in \cite{bender2018understanding},
PGNAS successfully finds better sub-networks which bring relatively 14\% - 23\% gain.
Evidently, the posterior distribution guided sampling is much more effective, which validates that our approach can learn a meaningful distribution for efficient architecture search. Besides, as can be viewed in the table, there is usually a huge performance gap between the architecture searched with predefined distribution with and without fine-tuning, which reveals the mismatching problems.

\begin{table}[]
\centering
     \begin{tabular}{c|ccccccc}
                 & 0.05K & 0.5k & 1.5k & 5.0k & 10k & 20k \\
                 \hline
        Error(\%)  & 2.17 & 2.06 & 2.06 & 2.04 & 1.98 & - \\
        $\Delta$GDays & 0.02 & 0.23 & 0.69 & 2.31 & 4.63 & 9.26 \\
        \hline
    \end{tabular}
\caption{Impact of the number of candidates on SupNet-MI.}
\label{ablationx_d}
\end{table}

Table. \ref{ablationx_b} discusses the weight prior $l$ in Eq. \ref{gumbelsft}. 
We find that a good $l$ usually makes the term $\sum_{l,i,s} \frac{(l_{l,k}^s)^{2}(1-p_l^s)}{2N}$ in Eq. \ref{deriv} fall into a commonly used weight decay range. So we choose $l$ by grid search. As shown in this table, the weight prior $l$ affects both error rate and model size. The higher the $l$ is, the smaller the size of parameters. We choose the one with the minimal error rate.  
\begin{table}[]
\centering
     \begin{tabular}{c|cccc}
                 & MI$^*$ & M & E & EI  \\
                 \hline
        Channel-level  & 0.44 & 0.18 & 0.33 & 0.29 \\
        Operation-level & 0.26 & 0.10 & 0.19 & 0.21 \\
        \hline
    \end{tabular}
\caption{Proportions of dropped channels and operations.}
\label{proportion}
\end{table}

Table. \ref{ablationx_c} shows the impact of temperature value $\tau$ in Eq. \ref{gumbelsft}. It shows that a smaller $\tau$ leads to a lower error, which is consistent with the analysis regarding to Eq. \ref{gumbelsft}. The corresponding fine-tuned result of our PGNAS also provides marginal improvement, which demonstrates the reliability of our PGNAS on sampling of both architecture and weights. 
\begin{table*}
\centering
\small
\begin{tabular}{c c c c c}
\hline
\hline
Method & Error(\%)(Top1/Top5) & GPUs Days & Params(M) & Search Method \\
\hline
\hline

NASNET-A \cite{zoph2018learning} & 26.0/8.4 & 1800 & 5.3 & RL \\






AmoebaNet-C \cite{real2018regularized} & \textbf{24.3}/7.6 & 3150 & 6.4 & evolution \\

PNAS \cite{liu2018progressive} & 25.8/8.1 & 225 & 5.1 & SMBO \\
\hline

BayesNAS ($\lambda^o_w=0.005$) \cite{zhou2019bayesnas} & 26.5/8.9 & 0.2 & 3.9 & gradient-based \\

FBNet-C \cite{Wu2018FBNetHE} & 25.1/- & 9 & 5.5 & gradient-based \\


DARTS \cite{liu2018darts} & 26.9/9.0 & 4 & 4.9 & gradient-based \\

SNAS \cite{xie2018snas} & 27.3/9.2 & 1.5 & 4.3 & gradient-based \\
\hline

One-Shot Top \cite{bender2018understanding} & 26.2/- & - & 6.8 & sampling-based \\

SinglePath \cite{guo2019single} & 25.3/- & 12 & - & sampling-based \\
\hline

PGNAS-D-121(ours) & 24.8/\textbf{7.5} & 26 & 6.6 & guided sampling \\

\hline
\hline
\end{tabular}
\caption{Comparison with other state-of-the-art results on ImageNet. Please note our model is directly searched on ImageNet.}
\label{imagenet}
\end{table*}

\begin{table}
\small
      \centering
        \begin{tabular}{c|ccc}
        Model  & Res50 & Inflated Res50 & PGNAS-R-50 \\
                 \hline
        Error  & 23.96\% & 22.93\% & \textbf{22.73}\% \\
        Params & 25.6M  & 44.0M & 26.0M \\
        \hline
        \end{tabular}
        \caption{Test results on ImageNet with inflated ResNet-50.}
        \label{res50}
\end{table}

We further evaluate the impact of number of samples in Table. \ref{ablationx_d}. The performance improves along with the increase of number of samples as well as the GPU days. Here we choose sampling 1500 architectures as a trade-off between the complexity and accuracy. Please also note that compared with other sampling-based NAS methods, our scheme achieves 2.17 \% error rate by sampling only 50 architectures with the assistance of the estimated a poseteriori distribution. It further reveals the fact that the estimated distribution provides essential information of the distribution of architectures and thus significantly facilitates the sampling process in terms of both efficiency and accuracy.

As discussed before, the correlated L2-like term in the derived objective function Eq. \ref{deriv} is not a conventional L2 regularization term. As demonstrated in Table. \ref{cifar100}, we observe severe performance drop with a constant weighted L2 term.

\textbf{Visualization.} We provide the visualization for the searched architecture of the best-performed PGNAS-MI$^*$ in supplementary materials. Given the initial super-network a multi-branched ResNet whose block structures are identical, PGNAS can still find diverse structure for each basic block.

\subsection{ImageNet}
We further evaluate our PGNAS on ImageNet with two super-networks based on ResNet50 \cite{he2016deep} and DenseNet121 \cite{huang2017densely}, respectively. Please find detailed settings in the supplementary material.  
Rather than transferring architectures searched on smaller dataset, the efficiency and flexibility of PGNAS enable us to directly search architectures on ImageNet within few days.


We first provide test results of our PGNAS on ImageNet in Table \ref{res50} using a relatively small search space by inflating ResNet50 without limiting the size of the model parameters. Hype-parameters and training process for the three models are identical for fair comparison.
It can be observed that PGNAS-R-50 outperforms the ResNet50 by 1.23\% with a similar size of parameters.
Table. \ref{imagenet} shows the comparison with the state-of-the-art results on ImageNet. Our method can search a very competitive architecture within 26 gpu days. Please note that we do not explicitly control the parameter size of the architectures searched by PGNAS because the goal of PGNAS is to find the architecture with the best accuracy. As can be viewed in Table. \ref{ablationx_b}, larger model does not necessarily generate better performance in our scheme. 

Please refer to supplementary materials for more performance evaluation and analysis.

\section{Discussions}
~~\textbf{Weight Sharing}. Weight sharing is a popular method adopted by one-shot models to greatly boost the efficiency of NAS. But it is not well understood why sharing weight is effective \cite{elsken2018neural,bender2018understanding}. In PGNAS, as discussed in subsection 2.2, we find that weight sharing can be viewed as a re-parametrization that enables us to estimate the posterior distribution through end-to-end optimization in our scheme.

\textbf{Network pruning}. Our method is a NAS method that conducts channel-level model selection, which is reminiscent of network pruning. We claim that the fundamental goal of NAS and network pruning is quite different. In fact, most of differentiable/one-shot NAS methods such as\cite{liu2018darts,cai2018proxylessnas,xie2018snas,Wu2018FBNetHE,bender2018understanding,guo2019single} start from a pre-defined cell/super-net and search for sub-architectures. Channel-level search only enables fine-grained architecture search (e.g., 'channel search space' in \cite{guo2019single}). Besides, it is often that all the channels of an operation are dropped, and PGNAS conducts operation-level selection in such cases. Please refer to Table. \ref{proportion} for the proportion of channels and operations that are not selected. Actually, in the re-parameterization process of Variational Inference that is used for approximating the posterior distribution, many schemes can be adopted. Dropout, which makes the sampling process resemble pruning, is just one of them. We could also use additive Gaussian noise for re-parameterization and the sampling will not involve pruning any more (section 3.2.2 \cite{gal2016uncertainty}).

\textbf{Limitations and Future Works}. One limitation of our PGNAS is that it can not explicitly choose the non-parametric operations such as pooling. Another one is that our PGNAS requires prior knowledge on architectures which is hard to achieve. Here we approaches the prior only by manually designed networks. 
So our future work may be 1) enabling selections on the non-parametric operations (e.g. assigning  a 1x1 convolution after each pooling operation as a surrogate to decide whether we need this pooling branch or not.) 2) investigating the robustness of our PGNAS to different prior architectures.


\section{Conclusion}
In this paper, we view the NAS problem from a Bayesian perspective and propose a new NAS approach, i.e. PGNAS, which converts NAS to a distribution construction problem. It explicitly approximates posterior distribution of network architecture and weights via network training to facilitate an more efficient search process in probability space. It also alleviates the mismatching problem between architecture and shared weights by sampling architecture-weights pair, which provides more reliable ranking results. The proposed PGNAS is efficiently optimized in an end-to-end way, and thus can be easily extended to other large-scale tasks. 

\section*{Acknowledgement}
This work was supported by the National Key R\&D Program of China under Grant 2017YFB1300201, the National Natural Science Foundation of China (NSFC) under Grants 61622211 and 61620106009 as well as the Fundamental Research Funds for the Central Universities under Grant WK2100100030.

{\small
\bibliographystyle{aaai}
\bibliography{egbib}
}

\end{document}